\documentclass[11pt,a4paper]{article}
\usepackage[hyperref]{naaclhlt2018}
\usepackage{times}
\usepackage{latexsym}
\usepackage{graphicx}
\graphicspath{{fig/}}
\usepackage{url}
\usepackage{multirow}
\usepackage{amsmath}
\usepackage{amssymb}
\usepackage{adjustbox}

\aclfinalcopy 
\def\aclpaperid{878} 

\newcommand{\newvec}[1]{\mathbf{#1}}
\DeclareMathOperator{\softmax}{\textbf{softmax}}

\title{TypeSQL: Knowledge-based Type-Aware Neural Text-to-SQL Generation}

\author{Tao Yu \\
  Yale University \\
  {\tt tao.yu@yale.edu} \\\And
  Zifan Li \\
  Yale University \\
  {\tt zifan.li@yale.edu} \\ \And
  Zilin Zhang \\
  Yale University \\
  {\tt zilin.zhang@yale.edu} \\ \AND
  Rui Zhang \\
  Yale University \\
  {\tt r.zhang@yale.edu} \\ \And
  Dragomir Radev \\
  Yale University \\
  {\tt dragomir.radev@yale.edu} \\}

\date{}

\begin{document}
\maketitle

\begin{abstract}
Interacting with relational databases through natural language helps users of any background easily query and analyze a vast amount of data. This requires a system that understands users' questions and converts them to SQL queries automatically.
In this paper we present a novel approach, \textsc{TypeSQL}, which views this problem as a slot filling task.
Additionally, \textsc{TypeSQL} utilizes type information to better understand rare entities and numbers in natural language questions.
We test this idea on the WikiSQL dataset and outperform the prior state-of-the-art by 5.5\% in much less time. We also show that accessing the content of databases can significantly improve the performance when users' queries are not well-formed. \textsc{TypeSQL} gets 82.6\% accuracy, a 17.5\% absolute improvement compared to the previous content-sensitive model.
\end{abstract}

\section{Introduction}

Building natural language interfaces to relational databases is an important and challenging problem \cite{li2014constructing,pasupat2015compositional,Yin15,Zhong2017,Yaghmazadeh17,Xu2017,Wang2017}.
It requires a system that is able to understand natural language questions and generate corresponding SQL queries.
In this paper, we consider the WikiSQL task proposed by \newcite{Zhong2017}, a large scale benchmark dataset for the text-to-SQL problem. Given a natural language question for a table and the table's schema, the system needs to produce a SQL query corresponding to the question.

\begin{figure*}[th!]
\centering
\includegraphics[width=\textwidth]{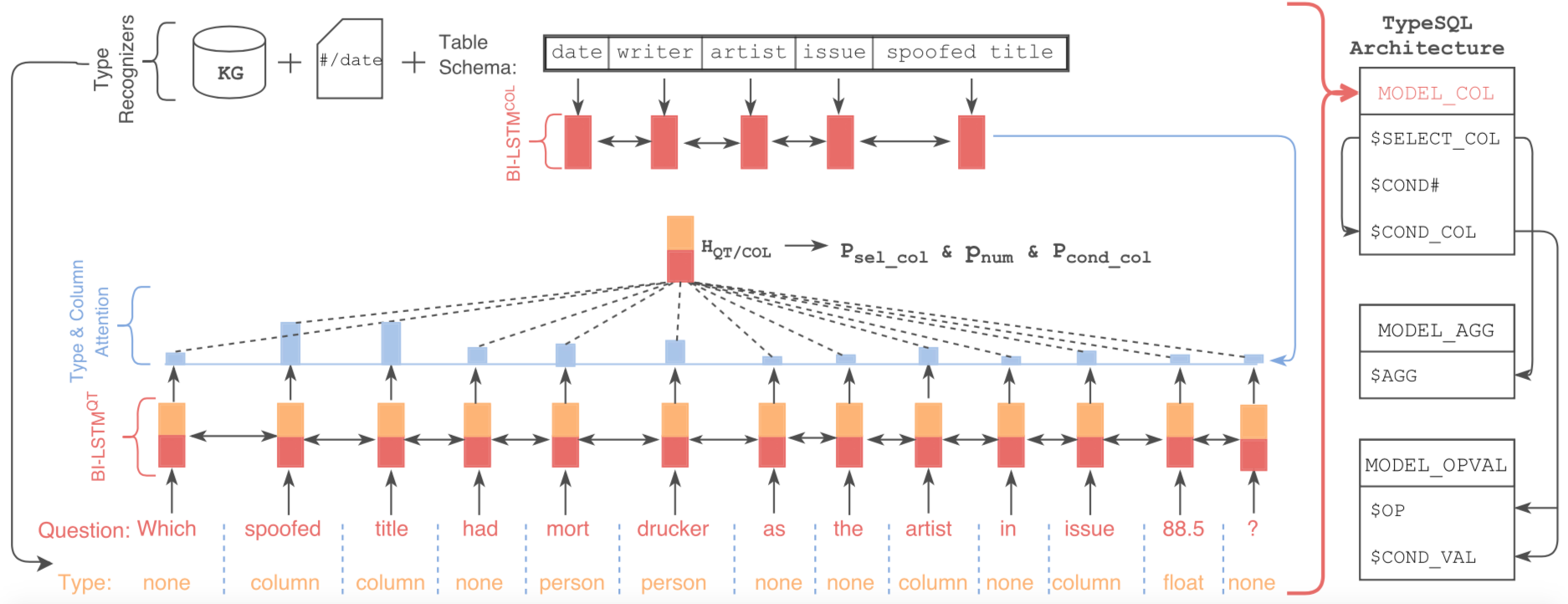}

\caption{\textsc{TypeSQL} consists of three slot-filling models on the right. We only show \textsc{model\_col} on the left for brevity. \textsc{model\_agg} and \textsc{model\_opval} have the similar pipelines.}
\label{fig:model}

\end{figure*}

We introduce a knowledge-based type-aware text-to-SQL generator, \textsc{TypeSQL}. Based on the prior state-of-the-art SQLNet \cite{Xu2017}, \textsc{TypeSQL} employs a sketch-based approach and views the task as a slot filling problem (Figure \ref{fig:sql_sketch}). By grouping different slots in a reasonable way and capturing relationships between attributes, \textsc{TypeSQL} outperforms SQLNet by about 3.5\% in half of the original training time.

Furthermore, natural language questions often contain rare entities and numbers specific to the underlying database. Some previous work \cite{agrawal03} already shows those words are crucial to many downstream tasks, such as infering column names and condition values in the SQL query. However, most of such key words lack accurate embeddings in popular pre-trained word embedding models. In order to solve this problem, \textsc{TypeSQL} assigns each word a type as an entity from either the knowledge graph, a column or a number. For example, for the question in Figure \ref{fig:model}, we label ``mort drucker" as \textsc{person} according to our knowledge graph; ``spoofed title," ``artist" and ``issue" as \textsc{column} since they are column names; and ``88.5" as \textsc{float}. Incorporating this type information, \textsc{TypeSQL} further improves the state-of-the-art performance by about another 2\% on the WikiSQL dataset, resulting in a final 5.5\% improvement in total. 

Moreover, most previous work assumes that user queries contain exact column names and entries. However, it is unrealistic that users always formulate their questions with exact column names and string entries in the table. To tackle this issue, when scaleability and privacy are not of a concern, the system needs to search databases to better understand what the user is querying.
Our content-sensitive model \textsc{TypeSQL + TC} gains roughly 9\% improvement compared to the content-insensitive model, and outperforms the previous content-sensitive model by 17.5\%.

\section{Related Work}
\label{sec:rel}

Semantic parsing maps natural language to meaningful executable programs. The programs could be a range of representations such as logic forms \cite{zelle96,Zettlemoyer05,wong07,Das10,Liang11,banarescu13,artzi13,Reddy14,Berant14,pasupat2015compositional}. Another area close to our task is code generation. This task parses natural language descriptions into a more general-purpose programming language such as Python \cite{Allamanis15,ling16, RabinovichSK17,Yin17}.

As a sub-task of semantic parsing, the text-to-SQL problem has been studied for decades \cite{warren1982efficient,popescu2003towards,popescu2004modern,li2006constructing,giordani2012translating,wang2017synthesizing}. The methods of the Database community \cite{li2014constructing, Yaghmazadeh17} involve more hand feature engineering and user interactions with the systems. In this work, we focus on recent neural network based approaches \cite{Yin15,Zhong2017,Xu2017,Wang2017,iyer17}. \newcite{dong16} introduce a sequence-to-sequence approach to converting text to logical forms. Most of previous work focus on specific table schemas, which means they use a single database in both train and test. Thus, they don't generalize to new databases.
\newcite{Zhong2017} publish the WikiSQL dataset and propose a sequence-to-sequence model with reinforcement learning to generate SQL queries. In the problem definition of the WikiSQL task, the databases in the test set do not appear in the train and development sets. Also, the task needs to take different table schemas into account. \newcite{Xu2017} further improve the results by using a SQL sketch based approach employing a sequence-to-set model.

\section{Methodology}
\label{sec:methods}

Like SQLNet, we employ a sketch-based approach and format the task as a slot filling problem. Figure \ref{fig:sql_sketch} shows the SQL sketch. Our model needs to predict all slots that begin with \texttt{\$} in Figure \ref{fig:sql_sketch}.

\begin{figure}[!t]
    \adjustbox{}{\textbf{SELECT} \texttt{\$AGG} \texttt{\$SELECT\_COL}}
    \adjustbox{}{\textbf{WHERE} \texttt{\$COND\_COL} \texttt{\$OP} \texttt{\$COND\_VAL}}
    \adjustbox{}{(\textbf{AND} \texttt{\$COND\_COL} \texttt{\$OP \texttt{\$COND\_VAL}})*}
    \caption{SQL Sketch. The tokens starting with ``\$" are slots to fill. ``*" indicates zero or more \textbf{AND} clauses.}
    \label{fig:sql_sketch}
\end{figure}

Figure \ref{fig:model} illustrates the architecture of \textsc{TypeSQL} on the right and a detailed overview of one of three main models \textsc{model\_col} on the left. We first preprocess question inputs by type recognition (Section \ref{sec:type}). Then we use two bi-directional LSTMs to encode words in the question with their types and the column names separately (Section \ref{sec:inputenc}). The output hidden states of LSTMs are then used to predict the values for the slots in the SQL sketch (Section \ref{sec:slot-filling}).

\subsection{Type Recognition for Input Preprocessing}
\label{sec:type}

In order to create one-to-one type input for each question, we, first, tokenize each question into $n$-grams of length 2 to 6, and use them to search over the table schema and label any column name appears in the question as \textsc{column}. Then, we assign numbers and dates in the question into four self-explanatory categories: \textsc{integer}, \textsc{float}, \textsc{date}, and \textsc{year}.
\urlstyle{same}
To identify named entities, we search for five types of entities: \textsc{person}, \textsc{place}, \textsc{country},  \textsc{organization}, and \textsc{sport}, on Freebase\footnote{\url{https://developers.google.com/freebase/}} using grams as keyword queries. The five categories cover a majority of entities in the dataset. Thus, we do not use other entity types provided by Freebase. Domain-specific knowledge graphs can be used for other applications.

In the case where the content of databases is available, we match words in the question with both the table schema and the content and labels of the columns as \textsc{column} and match the entry values as the corresponding column names. For example, the type in the Figure \ref{fig:model} would be [none, column, column, none, artist, artist, none, none, column, none, column, issue, none] in this case. Other parts in the Figure \ref{fig:model} keep the same as the content-insensitive approach.

\subsection{Input Encoder}
\label{sec:inputenc}

As shown in the Figure \ref{fig:model}, our input encoder consists of two bi-directional LSTMs, \textsc{bi-LSTM}$^{\textsc{QT}}$ and \textsc{bi-LSTM}$^{\textsc{COL}}$. To encode word and type pairs of the question, we concatenate embeddings of words and their corresponding types and input them to \textsc{bi-LSTM}$^{\textsc{QT}}$.
Then the output hidden states are $\newvec{H}_{\textsc{qt}}$ and $\newvec{H}_{\textsc{col}}$, respectively.

For encoding column names, SQLNet runs a bi-directional LSTM over each column name. We first average the embeddings of words in the column name. Then, we run a single \textsc{bi-LSTM}$^{\textsc{COL}}$ between column names. This encoding method improves the result by 1.5\% and cuts the training time by half. Even though the order of column names does not matter, we attribute this improvement to the fact that the LSTM can capture their occurrences and relationships.

\subsection{Slot-Filling Model}
\label{sec:slot-filling}

Next, we predict values for the slots in the SQL sketch. For the slots in Figure \ref{fig:sql_sketch}, SQLNet has a separate model for each of them which do not share their trainable parameters. This creates five models for the five slots and one model for \texttt{\$COND\#} (12 \textsc{bi-LSTM}s in total). However, since the predict procedures of \texttt{\$SELECT\_COL}, \texttt{\$COND\_COL}, and \texttt{\$COND\#} are similar, we combine them into a single model. Additionally, \texttt{\$COND\_COL} depends on the output of \texttt{\$SELECT\_COL}, which reduces errors of predicting the same column in these two slots \texttt{\$COND\_COL}  Moreover, we group \texttt{\$OP} and \texttt{\$COND\_VAL} together because both depend on the outputs of \texttt{\$COND\_COL}. Furthermore, we use one model for \texttt{\$AGG} because we notice that the \texttt{\$AGG} model converges much faster and suffers from overfitting when combined with other models.
Finally, \textsc{TypeSQL} consists of three models (Figure \ref{fig:model} right):
\begin{itemize}
    \item \textsc{model\_col} for \texttt{\$SELECT\_COL}, \texttt{\$COND\#} and \texttt{\$COND\_COL}
    \item \textsc{model\_agg} for \texttt{\$AGG}
    \item \textsc{model\_opval} for \texttt{\$OP} and \texttt{\$COND\_VAL}
\end{itemize}
where the parameters of \textsc{bi-LSTM}$^{\textsc{QT}}$ and \textsc{bi-LSTM}$^{\textsc{COL}}$ are shared in each model (6 \textsc{bi-LSTM}s in total).


Since all three models use the same way to compute the weighted question and type representation $\newvec{H}_{\textsc{qt/col}}$ using the column attention mechanism proposed in SQLNet, we first introduce the following step in all three models:

\begin{gather*}
\resizebox{0.7\hsize}{!}{$
\alpha_\textsc{qt/col} = \softmax(\newvec{H}_{\textsc{col}}\newvec{W}_{ct} \newvec{H}_{\textsc{qt}}^\top) $} \\
 \resizebox{0.5\hsize}{!}{$
\newvec{H}_{\textsc{qt/col}} = \alpha_\textsc{qt/col}\newvec{H}_{\textsc{qt}}
$}
\end{gather*}
where $\softmax$ applies the softmax operator over each row of the input matrix, $\mathbf{\alpha}_\textsc{qt/col}$ is a matrix of attention scores, and $\newvec{H}_{\textsc{qt/col}}$ is the weighted question and type representation. In our equations, we use $\newvec{W}$ and $\newvec{V}$ to represent all trainable parameter matrices and vectors, respectively.

\paragraph{\textsc{model\_col}-\texttt{\$SELECT\_COL}}
$\newvec{H}_{\textsc{qt/col}}$ is used to predict the column name in the \texttt{\$SELECT\_COL}:
\begin{gather*}
s = \newvec{V}^{sel} \textbf{tanh} (\newvec{W}_{c}^{sel} \newvec{H}_{\textsc{col}}^\top + \newvec{W}_{qt}^{sel} \newvec{H}_{\textsc{qt/col}}^\top) \\
P_{sel\_col}= \softmax(s)
\end{gather*}
\begin{table*}[!ht]
\small
\centering
\begin{tabular}{|l|l|l|l|l|l|l|}
\hline
\multirow{2}{*}{} & \multicolumn{3}{c|}{Dev}  & \multicolumn{3}{c|}{Test} \\ \cline{2-7} 
& Acc\textsubscript{lf} & Acc\textsubscript{qm} & Acc\textsubscript{ex} & Acc\textsubscript{lf} & Acc\textsubscript{qm} & Acc\textsubscript{ex} \\ \hline
\multicolumn{7}{|c|}{Content Insensitive} \\ \hline
\newcite{dong16}  & 23.3\%  & \multicolumn{1}{c|}{-} & 37.0\% & 23.4\% & \multicolumn{1}{c|}{-} & 35.9\% \\ \hline
Augmented Pointer Network \cite{Zhong2017}  & 44.1\%  & \multicolumn{1}{c|}{-} & 53.8\%  & 42.8\% & \multicolumn{1}{c|}{-} & 52.8\% \\ \hline
Seq2SQL \cite{Zhong2017}  & 49.5\%  & \multicolumn{1}{c|}{-} & 60.8\%  & 48.3\% & \multicolumn{1}{c|}{-} & 59.4\% \\ \hline
SQLNet \cite{Xu2017}  & \multicolumn{1}{c|}{-} & 63.2\%  & 69.8\% & \multicolumn{1}{c|}{-} & 61.3\%  & 68.0\%  \\ \hline
TypeSQL w/o type-awareness (ours)  & \multicolumn{1}{c|}{-} & 66.5\% & 72.8\% & \multicolumn{1}{c|}{-} & 64.9\% & 71.7\% \\ \hline
TypeSQL (ours)  & \multicolumn{1}{c|}{-} & \textbf{68.0\%} & \textbf{74.5\%} & \multicolumn{1}{c|}{-} & \textbf{66.7\%} & \textbf{73.5\%} \\ \hline
\multicolumn{7}{|c|}{Content Sensitive} \\ \hline
\newcite{Wang2017} & 59.6\% & \multicolumn{1}{c|}{-} & 65.2\%  & 59.5\% & \multicolumn{1}{c|}{-} & 65.1\% \\ \hline
TypeSQL+TC (ours) & \multicolumn{1}{c|}{-} & \textbf{79.2\%} & \textbf{85.5\%}  &   \multicolumn{1}{c|}{-} & \textbf{75.4\%} & \textbf{82.6\%} \\ \hline
\end{tabular}
\caption{Overall results on WikiSQL. Acc\textsubscript{lf}, Acc\textsubscript{qm}, and Acc\textsubscript{ex} denote the accuracies of exact string, canonical representation, and  execute result matches between the synthesized SQL with the ground truth respectively. The top six results are content-insensitive, which means only the question and table schema are used as inputs. The bottom two are content-sensitive, where the models use the question, the table schema, and the content of databases.}
\label{tb:ova_results}
\end{table*}

\begin{table*}[!ht]
\centering
\begin{tabular}{|l|l|l|l|l|l|l|}
\hline
\multirow{2}{*}{} & \multicolumn{3}{c|}{Dev} & \multicolumn{3}{c|}{Test} \\ \cline{2-7} 
                  & Acc\textsubscript{agg} & Acc\textsubscript{sel} & Acc\textsubscript{where} & Acc\textsubscript{agg} & Acc\textsubscript{sel} & Acc\textsubscript{where} \\ \hline
Seq2SQL \cite{Zhong2017} & 90.0\% & 89.6\% & 62.1\% & 90.1\% & 88.9\% & 60.2\% \\ \hline
SQLNet \cite{Xu2017} & 90.1\% & 91.5\% & 74.1\% & 90.3\% & 90.9\% & 71.9\% \\ \hline
TypeSQL (ours) & 90.3\% & \textbf{93.1\%} & \textbf{78.5\%} & 90.5\% & \textbf{92.2\%} & \textbf{77.8\%} \\ \hline
\hline
TypeSQL+TC (ours) & 90.3\% & \textbf{93.5\%} & \textbf{92.8\%} & 90.5\% & \textbf{92.1\%} & \textbf{87.9\%} \\ \hline
\end{tabular}
\caption{Breakdown results on WikiSQL. Acc\textsubscript{agg}, Acc\textsubscript{sel}, and Acc\textsubscript{where} are the accuracies of canonical representation matches on \textsc{aggregator}, \textsc{select column}, and \textsc{where} clauses between the synthesized SQL and the ground truth respectively.}
\label{tb:bkd_results}
\end{table*}

\paragraph{\textsc{model\_col}-\texttt{\$COND\#}}
Unlike SQLNet, we compute number of conditions in the \textsc{where} in a simpler way:
\begin{gather*}
 \resizebox{\hsize}{!}{$
P_{num}= \softmax \left(\newvec{V}^{num} \textbf{tanh} (\newvec{W}_{qt}^{num} \sum_{i} \newvec{H}^\top_{\textsc{qt/col}_i})\right)
$}
\end{gather*}
We set the maximum number of conditions to 4.

\paragraph{\textsc{model\_col}-\texttt{\$COND\_COL}}
We find that SQLNet often selects the same column name in the \texttt{\$COND\_COL} as \texttt{\$SELECT\_COL}, which is incorrect in most cases. To avoid this problem, we pass the weighted sum of question and type hidden states conditioned on the column chosen in \texttt{\$SELECT\_COL} $\newvec{H}_{\textsc{qt/scol}}$ (expended as the same shape of $\newvec{H}_{\textsc{qt/col}}$) to the prediction:

\begin{gather*}
 \resizebox{\hsize}{!}{$
c = \newvec{V}^{col} \textbf{tanh} (\newvec{W}_{c}^{col} \newvec{H}_{\textsc{col}}^\top + \newvec{W}_{qt}^{col} \newvec{H}_{\textsc{qt/col}}^\top + \newvec{W}_{qt}^{scol} \newvec{H}_{\textsc{qt/scol}}^\top)$ }\\
P_{cond\_col}= \softmax(c)
\end{gather*}

\paragraph{\textsc{model\_agg}-\texttt{\$AGG}}
Given the weighted sum of question and type hidden states conditioned on the column chosen in \texttt{\$SELECT\_COL} $\newvec{H}_{\textsc{qt/scol}}$, \texttt{\$AGG} is chosen from \{\textsc{null}, \textsc{max}, \textsc{min}, \textsc{count}, \textsc{sum}, \textsc{avg}\} in the same way as SQLNet:

\vspace{-3mm}
\begin{gather*}
\resizebox{0.85\hsize}{!}{$
P_{agg}= \softmax \left(\newvec{V}^{agg} \textbf{tanh} (\newvec{W}_{qt}^{agg} \newvec{H}_{\textsc{qt/scol}}^\top) \right)
$}
\end{gather*}

\paragraph{\textsc{model\_opval}-\texttt{\$OP}}
For each predicted condition column, we choose a \texttt{\$OP} from $\{=, >, <\}$ by:
\begin{gather*}
\resizebox{\hsize}{!}{$
P_{op} = \softmax \left(\newvec{W}_{t}^{op} \textbf{tanh} (\newvec{W}_{c}^{op} \newvec{H}_{\textsc{col}}^\top + \newvec{W}_{qt}^{op} \newvec{H}_{\textsc{qt/col}}^\top) \right)
$}
\end{gather*}

\paragraph{\textsc{model\_opval}-\texttt{\$COND\_VAL}} 
Then, we need to generate a substring from the question for each predicted column. As in SQLNet, a bi-directional LSTM is used for the encoder. It employs a pointer network \cite{Vinyals15} to compute the distribution of the next token in the decoder. In particular, the probability of selecting the $i$-th token $w_i$ in the natural language question as the next token in the substring is computed as:
\begin{gather*}
    \resizebox{\hsize}{!}{$v = \newvec{V}_{t}^{val}\textbf{tanh}(\newvec{W}_{qt}^{val}\newvec{H}^{i}_{\textsc{qt}} + \newvec{W}_{c}^{val}\newvec{H}_{\textsc{col}} + \newvec{W}_{h}^{val}\newvec{h})$}\\
    P_{cond\_val} = \softmax(v)
\end{gather*}
where $\newvec{h}$ is the hidden state of the previously generated token. The generation process continues until the $\langle$\texttt{END}$\rangle$ token is the most probable next token of the substring.

\section{Experiments}

\paragraph{Dataset}
We use the WikiSQL dataset \cite{Zhong2017}, a collection of 87,673 examples of questions, queries, and database tables built from 26,521 tables. It provides train/dev/test splits such that each table is only in one split. This requires model to generalize to not only new questions but new table schemas as well. 
\paragraph{Implementation Details}
We implement our model based on SQLNet \cite{Xu2017} in PyTorch \cite{paszke2017automatic}. We concatenate pre-trained Glove \cite{pennington14} and paraphrase \cite{Wieting17} embeddings. The dimensions and dropout rates of all hidden layers are set to 120 and 0.3 respectively. We use Adam \cite{Kingma15} with the default hyperparameters for optimization. The batch size is set to 64. The same loss functions in \cite{Xu2017} are used. Our code is available at \url{https://github.com/taoyds/typesql}.

\paragraph{Results and Discussion}
Table \ref{tb:ova_results} shows the main results on the WikiSQL task.
We compare our work with previous results using the three evaluation metrics used in \cite{Xu2017}.
Table \ref{tb:bkd_results} provides the breakdown results on \textsc{aggregation}, \textsc{selection}, and \textsc{where} clauses.

Without looking at the content of databases, our model outperforms the previous best work by 5.5\% on execute accuracy. According to Table \ref{tb:bkd_results}, \textsc{TypeSQL} improves the accuracy of \textsc{SELECT} by 1.3\% and \textsc{WHERE} clause by 5.9\%. By encoding column names and grouping model components in a simpler but reasonable way, \textsc{TypeSQL} achieves a much higher result on the most challenging sub-task \textsc{where} clause. Also, the further improvement of integrating word types shows that \textsc{TypeSQL} could encode the rare entities and numbers in a better way.

Also, if complete access to the database is allowed, \textsc{TypeSQL} can achieve 82.6\% on execute accuracy, and improves the performance of the previous content-aware system by 17.5\%. Although \cite{Zhong2017} enforced some limitations when creating the WikiSQL dataset, there are still many questions that do not have any column name and entity indicator. This makes generating the right SQLs without searching the database content in such cases impossible. This is not a critical problem for  WikiSQL but is so for most real-world tasks.

\section{Conclusion and Future Work}
\label{sec:conclusion}

We propose \textsc{TypeSQL} for text-to-SQL which views the problem as a slot filling task and uses type information to better understand rare entities and numbers in the input. \textsc{TypeSQL} can use the database content to better understand the user query if it is not well-formed. \textsc{TypeSQL} significantly improves upon the previous state-of-the-art on the WikiSQL dataset.

Although, unlike most of the previous work, the WikiSQL task requires model to generalize to new databases, the dataset does not cover some important SQL operators such as JOIN and GROUP BY. This limits the generalization of the task to other SQL components. In the future, we plan to advance this work by exploring other more complex datasets under the database-split setting. In this way, we can study the performance of a generalized model on a more realistic text-to-SQL task which includes many complex SQL and different databases.

\section*{Acknowledgement}
We thank Alexander Fabbri for his help in reviewing.

\bibliography{naaclhlt2018}

\begin{thebibliography}{34}
\expandafter\ifx\csname natexlab\endcsname\relax\def\natexlab#1{#1}\fi

\bibitem[{Agrawal and Srikant(2003)}]{agrawal03}
Rakesh Agrawal and Ramakrishnan Srikant. 2003.
\newblock Searching with numbers.
\newblock \emph{{IEEE} Trans. Knowl. Data Eng.}, 15(4):855--870.

\bibitem[{Allamanis et~al.(2015)Allamanis, Tarlow, Gordon, and
  Wei}]{Allamanis15}
Miltiadis Allamanis, Daniel Tarlow, Andrew~D. Gordon, and Yi~Wei. 2015.
\newblock Bimodal modelling of source code and natural language.
\newblock In \emph{{ICML}}, volume~37 of \emph{{JMLR} Workshop and Conference
  Proceedings}, pages 2123--2132. JMLR.org.

\bibitem[{Artzi and Zettlemoyer(2013)}]{artzi13}
Yoav Artzi and Luke Zettlemoyer. 2013.
\newblock Weakly supervised learning of semantic parsers for mapping
  instructions to actions.
\newblock \emph{Transactions of the Association forComputational Linguistics}.

\bibitem[{Banarescu et~al.(2013)Banarescu, Bonial, Cai, Georgescu, Griffitt,
  Hermjakob, Knight, Koehn, Palmer, and Schneider}]{banarescu13}
Laura Banarescu, Claire Bonial, Shu Cai, Madalina Georgescu, Kira Griffitt, Ulf
  Hermjakob, Kevin Knight, Philipp Koehn, Martha Palmer, and Nathan Schneider.
  2013.
\newblock Abstract meaning representation for sembanking.
\newblock In \emph{Proceedings of the 7th Linguistic Annotation Workshop and
  Interoperability with Discourse}.

\bibitem[{Berant and Liang(2014)}]{Berant14}
Jonathan Berant and Percy Liang. 2014.
\newblock Semantic parsing via paraphrasing.
\newblock In \emph{Proceedings of the 52nd Annual Meeting of the Association
  for Computational Linguistics (Volume 1: Long Papers)}, pages 1415--1425,
  Baltimore, Maryland. Association for Computational Linguistics.

\bibitem[{Das et~al.(2010)Das, Schneider, Chen, and Smith}]{Das10}
Dipanjan Das, Nathan Schneider, Desai Chen, and Noah~A. Smith. 2010.
\newblock Probabilistic frame-semantic parsing.
\newblock In \emph{NAACL}.

\bibitem[{Dong and Lapata(2016)}]{dong16}
Li~Dong and Mirella Lapata. 2016.
\newblock Language to logical form with neural attention.
\newblock In \emph{Proceedings of the 54th Annual Meeting of the Association
  for Computational Linguistics, {ACL} 2016, August 7-12, 2016, Berlin,
  Germany, Volume 1: Long Papers}.

\bibitem[{Giordani and Moschitti(2012)}]{giordani2012translating}
Alessandra Giordani and Alessandro Moschitti. 2012.
\newblock Translating questions to sql queries with generative parsers
  discriminatively reranked.
\newblock In \emph{COLING (Posters)}, pages 401--410.

\bibitem[{Iyer et~al.(2017)Iyer, Konstas, Cheung, Krishnamurthy, and
  Zettlemoyer}]{iyer17}
Srinivasan Iyer, Ioannis Konstas, Alvin Cheung, Jayant Krishnamurthy, and Luke
  Zettlemoyer. 2017.
\newblock Learning a neural semantic parser from user feedback.
\newblock \emph{CoRR}, abs/1704.08760.

\bibitem[{Kingma and Ba(2015)}]{Kingma15}
Diederik~P. Kingma and Jimmy Ba. 2015.
\newblock Adam: {A} method for stochastic optimization.
\newblock \emph{The 3rd International Conference for Learning Representations,
  San Diego}.

\bibitem[{Li and Jagadish(2014)}]{li2014constructing}
Fei Li and HV~Jagadish. 2014.
\newblock Constructing an interactive natural language interface for relational
  databases.
\newblock \emph{VLDB}.

\bibitem[{Li et~al.(2006)Li, Yang, and Jagadish}]{li2006constructing}
Yunyao Li, Huahai Yang, and HV~Jagadish. 2006.
\newblock Constructing a generic natural language interface for an xml
  database.
\newblock In \emph{EDBT}, volume 3896, pages 737--754. Springer.

\bibitem[{Liang et~al.(2011)Liang, Jordan, and Klein}]{Liang11}
P.~Liang, M.~I. Jordan, and D.~Klein. 2011.
\newblock Learning dependency-based compositional semantics.
\newblock In \emph{Association for Computational Linguistics (ACL)}, pages
  590--599.

\bibitem[{Ling et~al.(2016)Ling, Blunsom, Grefenstette, Hermann, Kocisk{\'{y}},
  Wang, and Senior}]{ling16}
Wang Ling, Phil Blunsom, Edward Grefenstette, Karl~Moritz Hermann, Tom{\'{a}}s
  Kocisk{\'{y}}, Fumin Wang, and Andrew Senior. 2016.
\newblock Latent predictor networks for code generation.
\newblock In \emph{{ACL} {(1)}}. The Association for Computer Linguistics.

\bibitem[{Pasupat and Liang(2015)}]{pasupat2015compositional}
Panupong Pasupat and Percy Liang. 2015.
\newblock Compositional semantic parsing on semi-structured tables.
\newblock In \emph{Proceedings of the 53rd Annual Meeting of the Association
  for Computational Linguistics and the 7th International Joint Conference on
  Natural Language Processing of the Asian Federation of Natural Language
  Processing, {ACL} 2015, July 26-31, 2015, Beijing, China, Volume 1: Long
  Papers}, pages 1470--1480.

\bibitem[{Paszke et~al.(2017)Paszke, Gross, Chintala, Chanan, Yang, DeVito,
  Lin, Desmaison, Antiga, and Lerer}]{paszke2017automatic}
Adam Paszke, Sam Gross, Soumith Chintala, Gregory Chanan, Edward Yang, Zachary
  DeVito, Zeming Lin, Alban Desmaison, Luca Antiga, and Adam Lerer. 2017.
\newblock Automatic differentiation in pytorch.
\newblock \emph{NIPS 2017 Workshop}.

\bibitem[{Pennington et~al.(2014)Pennington, Socher, and
  Manning}]{pennington14}
Jeffrey Pennington, Richard Socher, and Christopher~D. Manning. 2014.
\newblock Glove: Global vectors for word representation.
\newblock In \emph{{EMNLP}}, pages 1532--1543. {ACL}.

\bibitem[{Popescu et~al.(2004)Popescu, Armanasu, Etzioni, Ko, and
  Yates}]{popescu2004modern}
Ana-Maria Popescu, Alex Armanasu, Oren Etzioni, David Ko, and Alexander Yates.
  2004.
\newblock Modern natural language interfaces to databases: Composing
  statistical parsing with semantic tractability.
\newblock In \emph{Proceedings of the 20th international conference on
  Computational Linguistics}, page 141. Association for Computational
  Linguistics.

\bibitem[{Popescu et~al.(2003)Popescu, Etzioni, and Kautz}]{popescu2003towards}
Ana-Maria Popescu, Oren Etzioni, and Henry Kautz. 2003.
\newblock Towards a theory of natural language interfaces to databases.
\newblock In \emph{Proceedings of the 8th international conference on
  Intelligent user interfaces}, pages 149--157. ACM.

\bibitem[{Rabinovich et~al.(2017)Rabinovich, Stern, and Klein}]{RabinovichSK17}
Maxim Rabinovich, Mitchell Stern, and Dan Klein. 2017.
\newblock Abstract syntax networks for code generation and semantic parsing.
\newblock In \emph{{ACL} {(1)}}, pages 1139--1149. Association for
  Computational Linguistics.

\bibitem[{Reddy et~al.(2014)Reddy, Lapata, and Steedman}]{Reddy14}
Siva Reddy, Mirella Lapata, and Mark Steedman. 2014.
\newblock Large-scale semantic parsing without question-answer pairs.
\newblock \emph{Transactions of the Association for Computational Linguistics},
  2:377--392.

\bibitem[{Vinyals et~al.(2015)Vinyals, Fortunato, and Jaitly}]{Vinyals15}
Oriol Vinyals, Meire Fortunato, and Navdeep Jaitly. 2015.
\newblock Pointer networks.
\newblock In C.~Cortes, N.~D. Lawrence, D.~D. Lee, M.~Sugiyama, and R.~Garnett,
  editors, \emph{Advances in Neural Information Processing Systems 28}, pages
  2692--2700. Curran Associates, Inc.

\bibitem[{Wang et~al.(2017{\natexlab{a}})Wang, Brockschmidt, and
  Singh}]{Wang2017}
Chenglong Wang, Marc Brockschmidt, and Rishabh Singh. 2017{\natexlab{a}}.
\newblock Pointing out sql queries from text.
\newblock \emph{Technical Report}.

\bibitem[{Wang et~al.(2017{\natexlab{b}})Wang, Cheung, and
  Bodik}]{wang2017synthesizing}
Chenglong Wang, Alvin Cheung, and Rastislav Bodik. 2017{\natexlab{b}}.
\newblock Synthesizing highly expressive sql queries from input-output
  examples.
\newblock In \emph{Proceedings of the 38th ACM SIGPLAN Conference on
  Programming Language Design and Implementation}, pages 452--466. ACM.

\bibitem[{Warren and Pereira(1982)}]{warren1982efficient}
David~HD Warren and Fernando~CN Pereira. 1982.
\newblock An efficient easily adaptable system for interpreting natural
  language queries.
\newblock \emph{Computational Linguistics}, 8(3-4):110--122.

\bibitem[{Wieting and Gimpel(2017)}]{Wieting17}
John Wieting and Kevin Gimpel. 2017.
\newblock Pushing the limits of paraphrastic sentence embeddings with millions
  of machine translations.
\newblock \emph{arXiv preprint arXiv:1711.05732}.

\bibitem[{Wong and Mooney(2007)}]{wong07}
Yuk~Wah Wong and Raymond~J. Mooney. 2007.
\newblock Learning synchronous grammars for semantic parsing with lambda
  calculus.
\newblock In \emph{Proceedings of the 45th Annual Meeting of the Association
  for Computational Linguistics (ACL-2007)}, Prague, Czech Republic.

\bibitem[{Xu et~al.(2017)Xu, Liu, and Song}]{Xu2017}
Xiaojun Xu, Chang Liu, and Dawn Song. 2017.
\newblock Sqlnet: Generating structured queries from natural language without
  reinforcement learning.
\newblock \emph{arXiv preprint arXiv:1711.04436}.

\bibitem[{Yaghmazadeh et~al.(2017)Yaghmazadeh, Wang, Dillig, and
  Dillig}]{Yaghmazadeh17}
Navid Yaghmazadeh, Yuepeng Wang, Isil Dillig, and Thomas Dillig. 2017.
\newblock Sqlizer: Query synthesis from natural language.
\newblock \emph{Proc. ACM Program. Lang.}, 1(OOPSLA):63:1--63:26.

\bibitem[{Yin et~al.(2016)Yin, Lu, Li, and Kao}]{Yin15}
Pengcheng Yin, Zhengdong Lu, Hang Li, and Ben Kao. 2016.
\newblock Neural enquirer: Learning to query tables in natural language.
\newblock In \emph{Proceedings of the Twenty-Fifth International Joint
  Conference on Artificial Intelligence, {IJCAI} 2016, New York, NY, USA, 9-15
  July 2016}, pages 2308--2314.

\bibitem[{Yin and Neubig(2017)}]{Yin17}
Pengcheng Yin and Graham Neubig. 2017.
\newblock A syntactic neural model for general-purpose code generation.
\newblock In \emph{{ACL} {(1)}}, pages 440--450. Association for Computational
  Linguistics.

\bibitem[{Zelle and Mooney(1996)}]{zelle96}
John~M. Zelle and Raymond~J. Mooney. 1996.
\newblock Learning to parse database queries using inductive logic programming.
\newblock In \emph{AAAI/IAAI}, pages 1050--1055, Portland, OR. AAAI Press/MIT
  Press.

\bibitem[{Zettlemoyer and Collins(2005)}]{Zettlemoyer05}
Luke~S. Zettlemoyer and Michael Collins. 2005.
\newblock Learning to map sentences to logical form: Structured classification
  with probabilistic categorial grammars.
\newblock \emph{UAI}.

\bibitem[{Zhong et~al.(2017)Zhong, Xiong, and Socher}]{Zhong2017}
Victor Zhong, Caiming Xiong, and Richard Socher. 2017.
\newblock Seq2sql: Generating structured queries from natural language using
  reinforcement learning.
\newblock \emph{CoRR}, abs/1709.00103.

\end{thebibliography}
\bibliographystyle{acl_natbib}
\end{document}